\documentclass{article} 
\usepackage{iclr2025_conference,times}

\usepackage{amsmath,amsfonts,bm}

\def\eqref#1{equation~\ref{#1}}

\def\1{\bm{1}}

\DeclareMathAlphabet{\mathsfit}{\encodingdefault}{\sfdefault}{m}{sl}
\SetMathAlphabet{\mathsfit}{bold}{\encodingdefault}{\sfdefault}{bx}{n}

\usepackage{hyperref}
\usepackage{graphicx}
\usepackage{url}
\usepackage{listings}
\usepackage{booktabs}
\usepackage{subcaption}
\usepackage{pdfpages}

\title{TeleLoRA: Teleporting Model-Specific Alignment Across LLMs}

\author{Xiao Lin\thanks{Equal contributions}, Manoj Acharya\footnotemark[1], Anirban Roy \& Susmit Jha  \\
SRI International\\
Menlo Park, CA 94025, USA \\
\texttt{\{xiao.lin,manoj.acharya,anirban.roy,susmit.jha\}@sri.com} \\
}

\iclrfinalcopy 
\begin{document}

\maketitle

\begin{abstract}

Mitigating Trojans in Large Language Models (LLMs) is one of many tasks where alignment data is LLM specific, as different LLMs have different Trojan triggers and trigger behaviors to be removed. In this paper, we introduce \textbf{TeleLoRA} (\textbf{Tele}porting \textbf{Lo}w-\textbf{R}ank \textbf{A}daptation), a novel framework that synergizes model-specific alignment data across multiple LLMs to enable zero-shot Trojan mitigation on unseen LLMs without alignment data. TeleLoRA learns a unified generator of LoRA adapter weights by leveraging local activation information across multiple LLMs. This generator is designed to be permutation symmetric to generalize across models with different architectures and sizes. We optimize the model design for memory efficiency, making it feasible to learn with large-scale LLMs with minimal computational resources. Experiments on LLM Trojan mitigation benchmarks demonstrate that TeleLoRA effectively reduces attack success rates while preserving the benign performance of the models.

\end{abstract}

\section{Introduction}

“To each their own." It applies to Large Language Model (LLM) alignment too. Trojan mitigation~\cite{TrojAI,liu2024mitigating,chen2019deepinspect} is one such problem, where an LLM's weights have been modified to contain Trojans or backdoors that activate attacker-defined behaviors in response to a specific trigger string in the prompt. 
Mitigating these attacks such that the target LLM correctly rejects or ignores Trojaned prompts requires \textbf{model-specific alignment supervision} data as one LLM's Trojan behavior would not be present in another LLM. The diversity of triggers and the model-specific nature of the injected behaviors makes it a challenging problem.
Traditional approaches to Trojan mitigation often rely on expensive trigger reverse engineering and fine-tuning for each affected model~\cite{wang2019neural,gao2020backdoor}. This process is not scalable, especially with unseen LLMs with varying architectures, sizes, and Trojan behaviors, leaving new or proprietary LLMs vulnerable to such attacks.

\begin{figure}[t]
    \centering
    \includegraphics[width=0.80\linewidth]{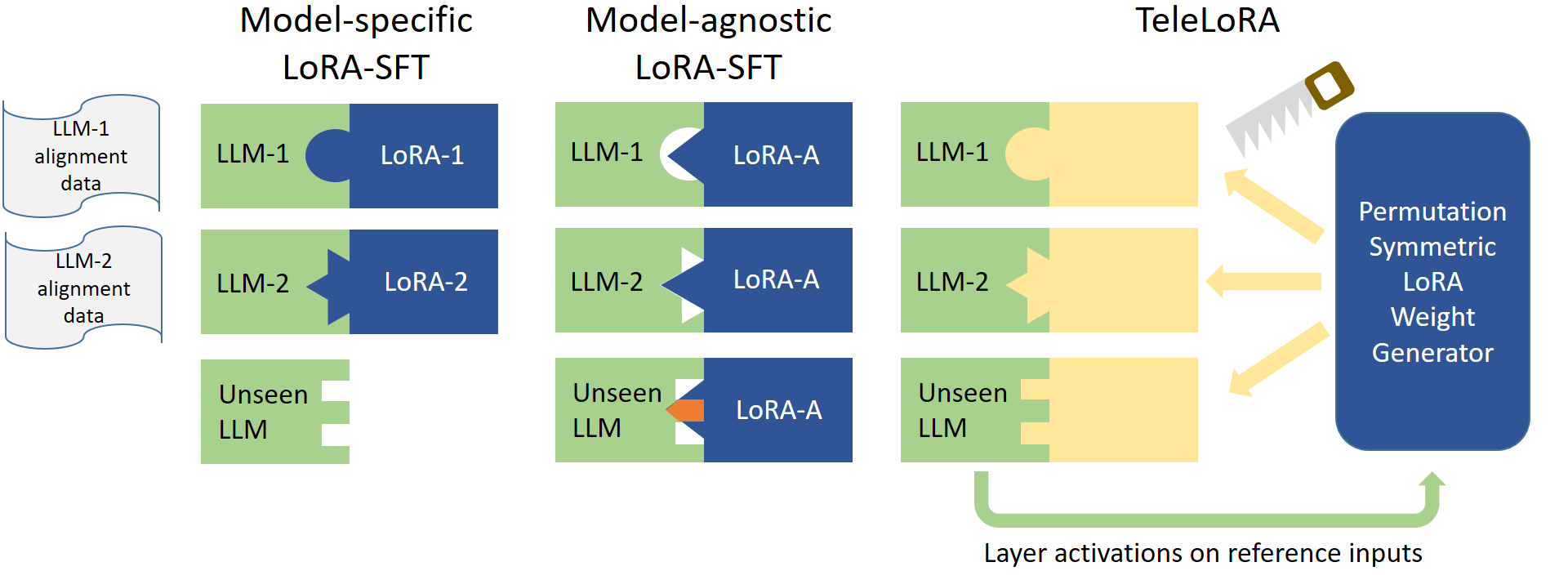}
    \caption{For model-specific alignment where different LLMs require different alignment supervision, TeleLoRA enables synergy over seen LLMs and zero-shot alignment on unseen LLMs by learning a unified generator of LoRA adapter weights across different LLMs. In contrast, model-specific adapters could not be learned on LLMs without alignment supervision. Model agnostic adapters learned on alignment supervision from other LLMs may not fit the current LLM. }
\label{fig:overview}
\end{figure}

In this work, we introduce \textbf{TeleLoRA} (\textbf{Tele}porting \textbf{Lo}w-\textbf{R}ank \textbf{A}daptation, Figure~\ref{fig:overview}), a novel framework that leverages weight-space learning~\cite{zhou2024permutation} for synergizing model-specific alignment over seen LLMs to perform zero-shot alignment on unseen LLMs by learning a unified generator of Low-Rank Adaptation (LoRA) adapter weights across different models given model activations on reference examples. 

Our approach makes TeleLoRA practical and scalable through:

\begin{enumerate}
    \item \textbf{Permutation symmetric model design:} We utilize a permutation symmetric neural network to efficiently generate LoRA adapter weights, significantly reducing the number of learnable parameters and computational complexity.
    \item \textbf{Memory Efficiency:} By sharing TeleLoRA modules across LLM layers and using techniques like gradient checkpointing, we minimize memory usage to enable training on multiple LLMs with limited GPU resources.
\end{enumerate}

We demonstrate effective cross-model alignment transfer with TeleLoRA on the IARPA-NIST LLM Trojan mitigation benchmark~\cite{TrojAI} and jailbreak mitigation~\cite{shen2024anything} tasks.

\section{Related works}

\paragraph{Trojan Mitigation in Neural Networks and LLMs.}

Trojan or backdoor attacks inject malicious behaviors into neural networks that activate upon specific trigger inputs~\cite{gu2017badnets, chen2017targeted}. In LLMs, the vast parameter space and complex token interactions complicates Trojan detection and mitigation. Fine-tuning-based mitigation approaches~\cite{liu2024mitigating} requires trigger specific knowledge for effective mitigation which is often hard to obtain. Anomaly detection on activations~\cite{yudin2023dubious} is not informed of Trojan trigger behavior and can be easily disabled by an attacker, limiting their performance. Our work differs by introducing a weight space modification framework that enables learnable zero-shot Trojan mitigation on unseen LLMs without requiring model-specific alignment data.

\paragraph{Weight-Space Learning and Hypernetworks in LLMs.}

Weight-space learning~\cite{von2019continual,zhou2024permutation} focuses on analysis and synthesis in the parameter space of neural networks, enabling performance analysis, parameter adaptation and sharing across tasks. Weight-space learning is facilitated by analyzing symmetry properties in neural architectures. 
Applying weight space learning to LLMs is challenging due to the sheer size of LLM weights. TeleLoRA makes this feasible by generating LoRA adapter weights which is orders of magnitude less than raw LLM weights, and through aggressive model sharing and memory optimizations. 

\section{Approach}

\begin{figure}[t]
    \centering
    \includegraphics[width=0.70\linewidth]{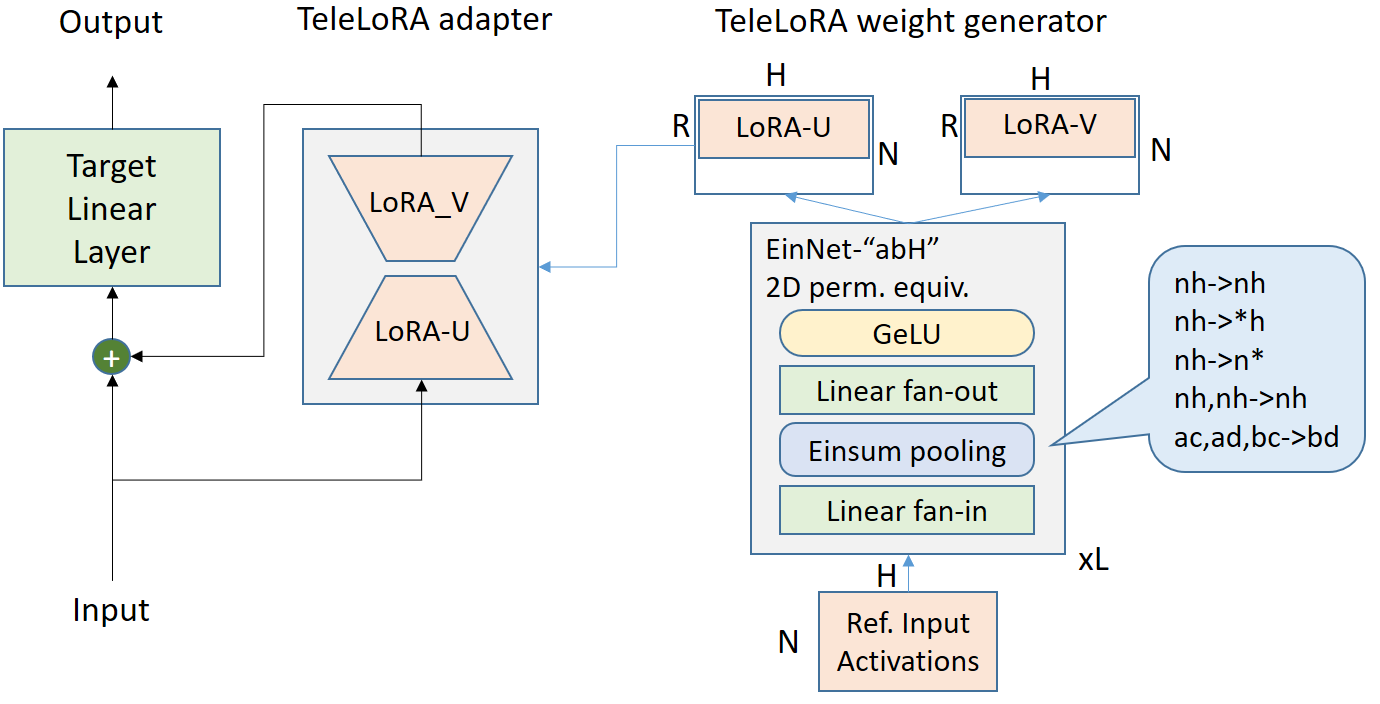}
    \caption{A TeleLoRA module on a linear layer uses local activations under reference inputs to predict weights of a multiplicative LoRA adapter for alignment. The network is invariant to the reference inputs (N), LoRA dimensions (R) and equivariant to neurons (H).}
\label{fig:telelora}
\end{figure}

\paragraph{TeleLoRA module for a linear layer.}
As shown in Figure~\ref{fig:telelora}, a TeleLoRA module uses local activation information to adapt the behavior of a linear layer. For a linear layer, given a set of sample input activations, TeleLoRA generates a LoRA weight adapter to the linear layer using a learnable permutation symmetric neural network. 

Specifically, given $X\in \mathbb{R}^{N\times H}$ -- a stack of $N$ activations with dimensionality $H$ -- TeleLoRA learns a neural network $U,V=f(X)$ to generate LoRA weights $U,V\in \mathbb{R}^{r\times H}$, that transform a linear layer $y=Wx+b$ into $y=W(I+V^TU)x+b$. 

The design of the TeleLoRA module follows two principles. First, the weight generation process is invariant to permutations of examples and LoRA ranks, and equivariant to neuron permutations. Second, memory efficiency is needed to accommodate backpropagation of large models such as LLMs alongside learning of TeleLoRA weight generation.

To achieve that, a TeleLoRA module processes input activations $X$ using a 2D permutation equivariant network along both sample and latent dimensions. The permutation equivariant network outputs two matrices $U_0, V_0 \in \mathbb{R}^{N\times H}$. We then randomly sample $r$ rows of $U_0$ and $V_0$ to create LoRA weights $U$ and $V$ respectively. In this way, we reuse the sample dimension $N$ as LoRA rank $r$ for greatly lowered memory cost while maintaining the required symmetry.

For the permutation symmetric backbone, we use an EinNet-ab backbone from prior work (see Appendix B for design and PyTorch code) which adds high order permutation symmetric operations such as $XX^TX$ to the popular EMLP~\cite{finzi2021practical} approach for better learning of high order matrix operations, please refer to the appendix for more details.

\paragraph{Cross-LLM alignment with TeleLoRA}

Given alignment data across multiple LLMs, we learn the TeleLoRA weight generator to optimize the alignment loss. If implemented naively, a single iteration involves forward-backward of both an LLM and the TeleLoRA weight generator which is memory intensive, so careful design of the weight generator is needed.

\textbf{Adapter sharing.} To enable compatibility across different LLMs, we learn one TeleLoRA module to every type of linear layer present in the general transformer architecture (e.g. qproj, kproj, vproj, etc.) and share the TeleLoRA module across different Transformer layers. With gradient checkpointing, module sharing allows weights of different LLM layers to be generated separately, reducing memory cost by a factor of 25 to 100. This is important for practical weight generation. 

\textbf{Gradient checkpointing.} To enable simultaneous meta learning over multiple memory intensive LLMs, we use gradient checkpointing to reduce memory cost. Each training iteration consists of 3 phases. 1) Randomly select an LLM, send to GPU and run forward pass on each TeleLoRA module to fill the LoRA weights. 2) Forward-backward on the LLM + LoRA to compute the gradient from the SFT loss to LoRA weights. 3) Forward-backward on each TeleLoRA module to backprop gradients from LoRA weights to TeleLoRA module parameters. Because LoRA weights are computed independently for each linear layer, the memory cost of step 1) and 3) are based on a single TeleLoRA module. The memory cost of step 2) is based on LoRA finetuning the largest LLM if inactive LLMs are unloaded off the GPU. In practice, this makes it possible to train TeleLoRA on multiple LLMs with up to 8B parameters on a single GPU within 24GB VRAM. 

\textbf{Reference activations.} Diversity of reference activations is important for generating high-quality adapter weights for different LLM layers. In practice, we find that N=50 reference activations from different text samples are often sufficient for layer-shared TeleLoRA to get close to per-layer LoRA.

\textbf{Multi-step inference.} When applying trained TeleLoRA modules, instead of generating LoRA adapters in a single step, we can also run TeleLoRA for multiple iterations to generate LoRA weights incrementally at small step sizes. We iteratively compute activations and update adapter weights with step size $\alpha$($=0.1$) over $K$($\in[3,10]$) steps. In practice, iterative weight generation often improves adapter quality on unseen models.

\section{Experiments}

We evaluate TeleLoRA against model-specific and model-agnostic LoRA on Trojan mitigation and jailbreak mitigation that require model-specific alignment.

\paragraph{Trojan Mitigation.}
The TrojAI LLM Trojan mitigation benchmark~\cite{TrojAI} tests Trojan mitigation on instruction-tuned LLMs with unknown Trojan triggers and behaviors. Two LLMs are provided for training, one clean (Gemma-2-2B) and one backdoored with a phrase trigger (Llama-3.1-8B). The sequestered test-set consists of 21 LLMs with undisclosed architectures and Trojan triggers. The evaluation metric is defined as \(\text{Fidelity} = \frac{\text{ASR}_{\text{pre-mitigation}} - \text{ASR}_{\text{post-mitigation}}}{\text{ASR}_{\text{pre-mitigation}}} \times \frac{\text{MMLU}_{\text{post-mitigation}}}{\text{MMLU}_{\text{pre-mitigation}}}\).

We train TeleLoRA across multiple LLMs to resist the provided Trojan triggers as well as generic jailbreak prompts from~\cite{shen2024anything} as data augmentation. For each LLM, we generate 1,000 alignment examples consisting of (poisoned question, clean answer) pairs. The poisoned question is created by pairing a question selected from the SQuADv2 dataset with a Trojan trigger when available or a random jailbreak prompt. The clean answer is generated from the clean base LLM given the clean SQuADv2 question. In addition to the 2 provided LLMs, we also include off-the-shelf Llama-3.2-1B and 3B Instruct, Qwen2.5-1.5B-Instruct, Gemma and Gemma2-2B-it and Phi-3.5-mini-Instruct for training, for a total of 8 LLMs. We compare TeleLoRA at 12 stacks, hidden size 16, $R=20$ against model-agnostic LoRA as the baseline, which finetunes the target LLM with all alignment examples for Trojan mitigation.

Results are shown in Table~\ref{table:TrojAI}. TeleLoRA achieves the best-in-class mitigation ASR on poisoned LLMs, with a small penalty on MMLU performance and overall fidelity score close to the best method. Compared to LoRA, TeleLoRA shows a significant improvement in the Fidelity metric, primarily due to its much better Trojan mitigation ASR.

\begin{table}[t]
\centering
\begin{subtable}{0.60\textwidth}
\centering
\begin{tabular}{l|ccc|c}
\toprule
Method & PSN & PSN & Clean & Fidelity \\
       & ASR $\downarrow$ & MMLU $\uparrow$ & MMLU $\uparrow$ & $\uparrow$ \\
\midrule
Perspecta* & 0.66 & 0.58 & 0.58 & 0.52  \\
LoRA      & 0.72 & 0.56 & 0.56 & 0.45  \\
\midrule
TeleLoRA  & 0.59 & 0.53 & 0.51 & 0.51 \\
\bottomrule
\end{tabular}
\caption{Results on the holdout splits of the TrojAI \texttt{mitigation-llm-instruct-oct2024} dataset. (*Perspecta is a participant in the TrojAI challenge.)}
\label{table:TrojAI}
\end{subtable}
\hfill
\begin{subtable}{0.38\textwidth}
\centering
\begin{tabular}{l|c}
\toprule
Method & Synergy  \\
       & Avg PPL Seen $\downarrow$  \\
\midrule
LoRA-specific  & 11.31    \\
LoRA-agnostic  & 11.86  \\
\midrule
TeleLoRA       & 10.96   \\
\bottomrule
\end{tabular}
\caption{Ablation study on synergy (PPL seen) for jailbreak mitigation.}
\label{table:ablation}
\end{subtable}
\label{table:combined}
\end{table}

\paragraph{Jailbreak Mitigation.} For ablation studies, we study whether TeleLoRA synergizes non-overlapping jailbreaks across different LLMs. We select 8 diverse LLMs (Appendix A for full list). We assign 5 jailbreaks in~\cite{shen2024anything} to each LLM with 100 alignment examples per jailbreak for training. To evaluate synergy, we study whether LoRA mitigates all 40 jailbreaks across all LLMs, with perplexity on 100 unseen alignment examples as the metric. We compare TeleLoRA against model-specific LoRA which trains only on the jailbreaks assigned, and model-agnostic LoRA which trains on all alignment examples which may not match the current LLM.

Results are shown in Table~\ref{table:ablation} and are also publicly \footnote{\url{https://pages.nist.gov/trojai/\#mitigation-llm-instruct-oct2024:\~:text=Best\%20Results\%20based\%20on\%20Fidelity}} accessible. TeleLoRA achieves lower perplexity than model specific and agnostic LoRA methods, which indicates that the synergy effect from TeleLoRA is stronger than the sacrifices made to achieve cross LLM generalization.

\section{Conclusion}
By leveraging a permutation symmetric neural network that efficiently generates LoRA adapter weights based on local activations, we show TeleLoRA can effectively synergize alignment data from multiple LLMs and enable zero-shot adaptation on new, unseen LLMs for significantly reduced attack success rates while maintaining benign model performance.


\section*{Acknowledgments}
This material is based upon work supported by the Intelligence Advanced Research Projects Agency (IARPA) and Army Research Office (ARO) under Contract No. W911NF-20-C-0038. Any opinions, findings and conclusions or recommendations expressed in this material are those of the author(s) and do not necessarily reflect the views of the Intelligence Advanced Research Projects Agency (IARPA) and Army Research Office (ARO).

\bibliography{./iclr2025_conference}
\bibliographystyle{iclr2025_conference}



\section*{Supplementary material}

## A LLMs used for jailbreak study

Seen LLMs: 1) google/gemma-2-2b-it, 2) meta-llama/Llama-3.1-8B-Instruct, 3) google/gemma-2b-it, 4) meta-llama/Llama-3.2-3B-Instruct, 5) meta-llama/Llama-3.2-1B-Instruct, 6) microsoft/Phi-3.5-mini-instruct, 7) Qwen/Qwen2.5-1.5B-Instruct, 8) Qwen/Qwen2.5-3B-Instruct.

## B Appendix: Permutation Symmetric Neural Networks

Our approach analyzes the parameter sharing pattern induced by permutation symmetry in the Taylor series of a function, leveraging the null-space method from Finzi et al. (2021). We find that einsum operations are the fundamental building blocks needed for building high-order permutation invariant networks. We propose a network architecture that interleaves MLPs with einsum pooling operations for building generic permutation symmetric neural networks. Finally, we demonstrate the flexibility of permutation symmetric neural networks in the design of Trojan detection classifiers, ingesting weight matrices, neural activations and confusion matrices as features for Trojan detection.

We first introduce the theoretical approach of using Taylor series to parameterize permutation symmetric functions. And then, we introduce our design of practical permutation symmetric neural networks guided by theory. Finally, we introduce how such a network can be applied to Trojan detection.

### B.1 Paramterizing Symmetric Functions in Taylor Series

We introduce a simple yet general Taylor series-based view necessary for studying complex symmetry patterns. Specifically, we enable efficient universal learners – that can represent any such invariant or equivariant functions. The general idea is similar to parallel works Equivariant Multilayer Perceptrons (EMLP) Finzi et al. (2021), universal equivariant MLPs Ravanbakhsh (2020), Equivariant Polynomial layersPuny et al. (2023) as well as many others, but our approach adds further simplifications to focus on parameter count and compute efficiency.

Given the desired input-output shapes and symmetry constraints, we would proceed with the following steps: 1) Express a general function that matches the input-output shapes in Taylor series form. 2) Map the symmetry constraints into equations about the Taylor series coefficients. 3) Solve the equations for free parameters and the parameter sharing patterns, and parameterize the function using the free parameters. 4) Simplify the parameterization for efficient computation.

Let us use 1D permutation invariance as an example.

Let's say we want to make a function

$$y = f\left(\begin{bmatrix} x_0 & x_1 & x_2 \end{bmatrix}\right)$$

invariant to permutation of $x_0, x_1, x_2$.

Consider the Taylor series

$$\begin{aligned} &f\left(\begin{bmatrix} x_0 & x_1 & x_2 \end{bmatrix}\right) \\ =& a + \begin{bmatrix} b_0 & b_1 & b_2 \end{bmatrix} \begin{bmatrix} x_0 \\ x_1 \\ x_2 \end{bmatrix} + \begin{bmatrix} x_0 & x_1 & x_2 \end{bmatrix} \begin{bmatrix} c_{00} & c_{01} & c_{02} \\ c_{10} & c_{11} & c_{12} \\ c_{20} & c_{21} & c_{22} \end{bmatrix} \begin{bmatrix} x_0 \\ x_1 \\ x_2 \end{bmatrix} + \ldots \end{aligned}$$

Since we want $f(\cdot)$ to be invariant to any permutation matrix $P$, the invariant constraint says

$$f\left(\begin{bmatrix} x_0 \\ x_1 \\ x_2 \end{bmatrix}\right) = f\left(\begin{bmatrix} & P & \end{bmatrix} \begin{bmatrix} x_0 \\ x_1 \\ x_2 \end{bmatrix}\right)$$

For our Taylor series form, because of the uniqueness of Taylor series, all order-k coefficients on the left hand side need to match the corresponding order-k coefficients on the right hand side. That is for any permutation matrix $P$ we have

$$a = a$$

$$\begin{bmatrix} b_0 & b_1 & b_2 \end{bmatrix} = \begin{bmatrix} b_0 & b_1 & b_2 \end{bmatrix} \begin{bmatrix} P \end{bmatrix}$$

$$\begin{bmatrix} c_{00} & c_{01} & c_{02} \\ c_{10} & c_{11} & c_{12} \\ c_{20} & c_{21} & c_{22} \end{bmatrix} = \begin{bmatrix} P^T \end{bmatrix} \begin{bmatrix} c_{00} & c_{01} & c_{02} \\ c_{10} & c_{11} & c_{12} \\ c_{20} & c_{21} & c_{22} \end{bmatrix} \begin{bmatrix} P \end{bmatrix}$$

$$\ldots$$

These equations are all linear equations about coefficients $a$, $b_i$ and $c_{ij}$. So we can just enumerate all $P$ to get all the equations, and then solve them. For $b_i$ for example, enumerating different permutations $P$ would give

$$\begin{bmatrix} b_0 \\ b_1 \\ b_2 \end{bmatrix} = \begin{bmatrix} b_0 \\ b_2 \\ b_1 \end{bmatrix} = \begin{bmatrix} b_1 \\ b_0 \\ b_2 \end{bmatrix} = \begin{bmatrix} b_1 \\ b_2 \\ b_0 \end{bmatrix} = \begin{bmatrix} b_2 \\ b_0 \\ b_1 \end{bmatrix} = \begin{bmatrix} b_2 \\ b_1 \\ b_0 \end{bmatrix}$$

That is more than enough to say $b_0 = b_1 = b_2$. So the order-1 term has only 1 degree of freedom.

For $c_i$ there are more equations, but it turns out that solving the equations across all permutations would yield $c_{00} = c_{11} = c_{22}$ and $c_{01} = c_{10} = c_{10} = c_{12} = c_{20} = c_{21}$. So the order 2 term has 2 degrees of freedom, one for the diagonal and one for everywhere else.

Applying what we have learned, we can now write

$$y = f(\begin{bmatrix} x_0 & x_1 & x_2 \end{bmatrix})$$
$$= a + \begin{bmatrix} b & b & b \end{bmatrix} \begin{bmatrix} x_0 \\ x_1 \\ x_2 \end{bmatrix} + \begin{bmatrix} x_0 & x_1 & x_2 \end{bmatrix} \begin{bmatrix} c_0 & c_1 & c_1 \\ c_1 & c_0 & c_1 \\ c_1 & c_1 & c_0 \end{bmatrix} \begin{bmatrix} x_0 \\ x_1 \\ x_2 \end{bmatrix} + \ldots$$

For a total of 4 free parameters up to order 2, instead of 13 free parameters without the invariance constraint. More generally, for $N$ inputs, we still only need 4 parameters to express any permutation invariant function, whereas a non-invariant function needs $N^2 + N + 1$ parameters. In practice, parameterizing with symmetry often **reduces parameter count** exponentially.

We can further simplify by focusing on the free parameters

$$y = f(\begin{bmatrix} x_0 & x_1 & x_2 \end{bmatrix})$$
$$= a + b \sum_i x_i + (c_0 - c_1) \sum_i x_i^2 + c_1 \sum_i \sum_j x_i x_j + \ldots$$
$$= a + b \sum_i x_i + (c_0 - c_1) \sum_i x_i^2 + c_1 (\sum_i x_i)^2 + \ldots$$

An important effect of this simplification is **reduced compute**. It now requires $O(N)$ compute for $N$ inputs instead of $O(N^2)$ for order-2.

In math terms, the number of free parameters is the dimensionality of the null space of the symmetry equations. The free parameters can be numerically solved from the basis of this null space. But note that as the basis is often not unique, numerical solutions can vary by a linear combination and therefore may not be compute-optimal, so further simplification is still needed.

3

Although we didn't unroll order-3 and higher terms because they are difficult to visualize, they can still be analyzed with the same approach. Just imagine a cube or a hypercube of parameters, apply the symmetry transformations simultaneously along all dimensions and solve for the parameter sharing pattern.

In this section, we have learned that

1) Symmetry constraints reduce the number of free parameters.

2) A Taylor-series technique can be used to parameterize symmetric functions.

3) Different symmetries can have different impacts on degrees of freedom.

4) Certain parameterizations can reduce compute exponentially.

5) Parameterization of equivariant functions are tied to parameterization of invariant functions

6) Permutation invariant and equivariant functions can be parameterized solely using tensor contraction terms.

A Taylor series parameterization is sound in theory. In practice however, functions compound and high order interactions are common. Taylor series often provides too little relevant capacity and too much irrelevant capacity to be useful. Engineering is key in the journey to create universal learners of symmetric functions. In the next section, we'll focus on permutation symmetry and design a family of practical invariant and equivariant networks for various flavors of permutation symmetry.

### B.2 Engineering a Network with Permutation Symmetry

From matrices to sets to symbolic processing, permutation symmetry is found in many problems and requires extra attention during modeling. When handled properly however, permutation symmetry is also a blessing. As we have learned in the previous section, if parameterized properly, permutation symmetry has the potential to exponentially reduce parameter count and compute for highly efficient learning. At the other end of the spectrum, reciting the success recipe of deep learning, we can scale the latent dimension and stack equivariant layers to create exponentially more expressive networks at the same parameter count and compute as a regular network.

Devil's in the details, in this section we'll walk through the design of permutation symmetric neural networks for various types of permutation symmetry.

There are many places where you'll see permutation symmetry and they often come in different forms. So we'll first start from a summary of common types of permutation symmetry, and then the design of permutation equivariant layers. Permutation symmetry turns out to be closely tied to tensor contractions. That would allow us to synthesize efficient high-order permutation equivariant layers automatically in a procedural manner. Finally, we'll discuss further optimizations that helps practical implementation.

**Common types of permutation symmetry.** In the following table we analyze a few common problems by their type of permutation symmetry.

To aid discussions, we use a custom notation to describe the specific type of permutation symmetry, to capture both the input shape and the unique permutation axes. We have multiple dimensions and joint permutations whose symmetry notations are straight forward. There's also a dependency aspect, which captures a concept of "list of different lists". A fully independent batch dimension Z and a non-symmetric latent dimension H may be added optionally.

**Creating a permutation equivariant layer with einsum pooling.** Across all types of permutation symmetry, as we learned in Section I through Taylor series, it turns out that tensor contractions are are all you need for parameterizing permutation invariant and equivariant layers, which can then be stacked into a deep network.

Intuitively, tensor contractions like

$$Y_{ij} = \sum_k \sum_l X_{ik} X_{lk} X_{lj}$$

4

| Problem | Illustration | Symmetry type | Dependency |
|---|---|---|---|
| Predicting important weights to keep in a linear layer while pruning. Equivariant to neuron permutations along input 'a' and output 'b'. | 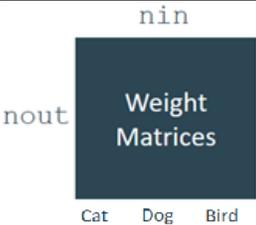 | ab | |
| Group classes by common misclassifications given confusion matrices. Equivariant to class permutations 'a' which simultaneously permutes both axes. | 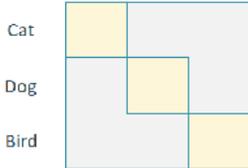 | aa | |
| Predicting the winner given players' hands. Equivariant to permutation of player ID a. Invariant to permutation of suits c. Invariant to card reordering b for each player independently. No symmetry against card number H. | [♡A,♦A,♣K,♡K,♦K]<br>[♠A,♠8,♠5,♠3,♠2]<br>[♦6,♦5,♦4,♦3,♦2]<br>[♣A,♠K,♡Q,♦J,♡10] | abcH | a→b |

Table 1: Common permutation symmetries and their symmetry type notations.

create a new tensor that has the same shape as the input while summing over unused dimensions. They achieve a permutation equivariant effect. And tensor contractions like

$$y = \sum_i \sum_j \sum_k \sum_l X_{ik} X_{lk} X_{lj}$$

that sum over all dimensions achieve a permutation invariant effect.

As the math equations can get quite long, we will use the einsum notation[1] which represents a tensor contraction using the indices involved. It is widely used across deep learning frameworks to denote tensor contractions. For example,

```
Y=einsum('Zik,Zlk,Zlj->Zij',X,X,X)
y=einsum('Zik,Zlk,Zlj->Z',X,X,X)
```

Here a batch dimension Z is added to make sure the right hand side is not empty.

**How to create an equivariant layer given permutation symmetry type from tensor contractions?** Our answer is two fully connected layers with a pooling layer in between.

Let us use a 1D + latent aH-type equivariant constraint as an example to illustrate the design.

The Taylor Series parameterization up to order 2 is

```
Y_abH=einsum('a->ba',a_H)
     +einsum('ab,ca->cb',b0_HH,X_aH)
     +einsum('ab,ca->db',b1_HH,X_aH)
     +einsum('abc,da,db->dc',c0_HHH,X_aH,X_aH)
     +einsum('abc,da,db->ec',c1_HHH,X_aH,X_aH)
     +einsum('abc,da,eb->dc',c2_HHH,X_aH,X_aH)
     +einsum('abc,da,eb->fc',c3_HHH,X_aH,X_aH)
     +...
```

---
[1] https://numpy.org/doc/stable/reference/generated/numpy.einsum.html

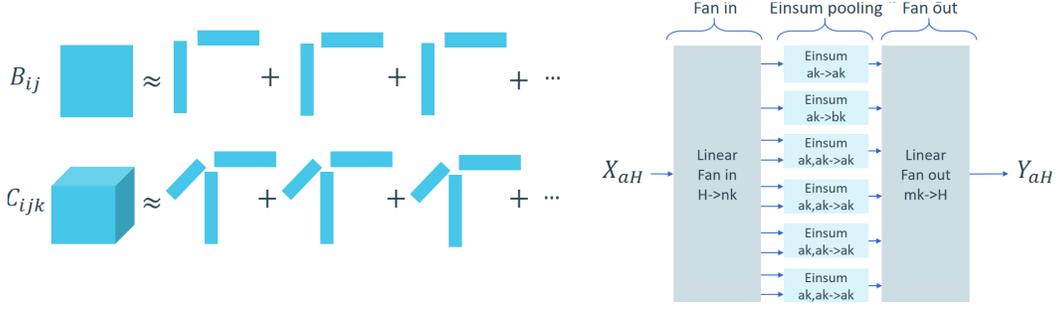

Figure 1: Left: Low-rank decomposition of high order coefficients. Right: An equivariant einsum layer with two linear layers an einsum pooling in between, which is the result of moving low-rank coefficients into linear fan-in and fan-out layers.

We can immediately see that the order-1 terms have $H^2$ parameters and order-2 terms have $H^3$ parameters, which would naturally need a low-rank($= K$) treatment, such as

```
Y_abH=einsum('a->ba',a_H)
     +einsum('ka,kb,ca->cb',b0U_KH,b0V_KH,X_aH)
     +einsum('ka,kb,ca->db',b1U_KH,b1V_KH,X_aH)
     +einsum('ka,kb,kc,da,db->dc',c0U,c0V,c0W,X_aH,X_aH)
     +einsum('ka,kb,kc,da,db->ec',c1U,c1V,c1W,X_aH,X_aH)
     +einsum('ka,kb,kc,da,eb->dc',c2U,c2V,c2W,X_aH,X_aH)
     +einsum('ka,kb,kc,da,eb->fc',c3U,c3V,c3W,X_aH,X_aH)
     +...
```

We can move the order-0 parameters, as well as $U, V, W$ matrices into fully connected layers along $H$ that perform input preprocessing and output postprocessing. So the end result is two linear layers with pooling in between, and for pooling we need

```
Y_abH_0V=einsum('ck->ck',X_aH_0U)
Y_abH_1V=einsum('ck->dk',X_aH_1U)
Y_abH_0W=einsum('dk,dk->dk',X_aH_0U,X_aH_0V)
Y_abH_1W=einsum('dk,dk->ek',X_aH_1U,X_aH_1V)
Y_abH_2W=einsum('dk,ek->dk',X_aH_2U,X_aH_2V)
Y_abH_3W=einsum('dk,ek->fk',X_aH_3U,X_aH_3V)
...
```

Notice that dk,ek−>fk can be composed with ck−>dk for each operand individually, and then combine using dk,dk−>dk. As we can stack more layers, not all pooling operations are needed and less pooling operations would reduce network complexity. In fact, this might be a good point to step back and ask: **Given equivariance type, e.g. aH, how can we identify the minimum yet sufficient set of pooling operations?**

The following recipe might be helpful for designing pooling operations given equivariance type in practice:

1) Enumerate all valid and unique einsum operations up to order-k that are compatible with the given equivariance type. For example einsum('ab,bc−>ac',X_ab,X_ab) is compatible with aa-type equivariance, but not compatible with ab-type equivariance. Also notice that ba,ac−>bc is the just a renaming of ab,bc−>ac. There is a graph homomorphism problem under the hood for listing unique einsum operations and interested readers can dig deeper.

2) Filter einsum operations based on dependency requirement of the given equivariance type. For example einsum('ab,cb−>cb',X_ab,X_ab) satisfy b−>a dependency but does not satisfy a−>b dependency for ab-type equivariance.

| Symmetry type | Order | Pooling operation(s) |
|---|---|---|
| aH | 1 | aH→aH, aH→bH |
|  | 2 | aH,aH→aH |
|  | 3+ | No need |
| abH | 1 | abH→abH, abH→cbH, abH→acH |
|  | 2 | abH,abH→abH |
|  | 3 | abH,cbH,cdH→adH |
|  | 4 | No need |
|  | 5 | abH,acH,dbH,dcH,deH→aeH, abH,acH,dbH,dcH,ecH→ebH |
| aaH | 1 | abH→abH,aaH→aaH,abH→baH,abH→cbH |
|  | 2 | abH,bcH→acH,abH,abH→abH |
|  | 3+ | No need |

Table 2: Common equivariance types and their required einsum pooling operations.

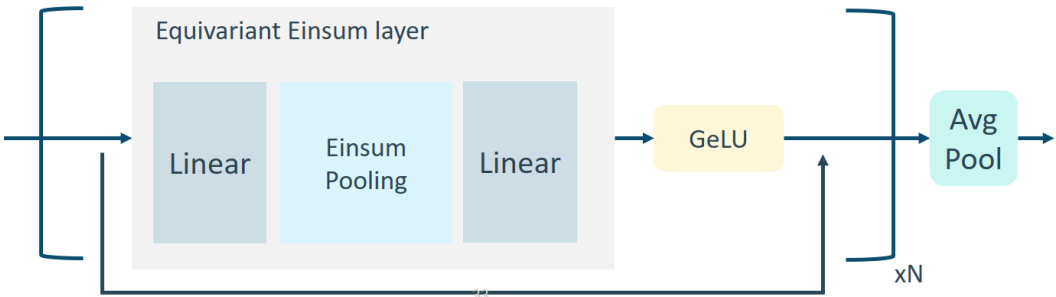

Figure 2: Stacking multiple equivariant layers to create a practical high capacity network.

3) Filter out order-2+ "breakable" operations that can be divided into two lower order terms with a simple pointwise multiplication. For example ab,cb,cd,ad−>ad can be divided into ab,cb,cd−>ad and ad−>ad which can the be put together with ab,ab−>ab, so it is not necessary as long as the lower order terms exist.

4) Normalize the rotation of input/output terms. For example for aa-type equivariance, ab,cb−>ca is not necessary because it can be achieved with ab,bc−>ac, through applying rotations ab−>ba on the input and output.

5) Remove order-2+ operations that expand new dimensions in the output term. For example ab,bc−>ad is redundant because it can be achieved through ab,bc−>ac followed by a dimension expansion operation ab−>ac.

An algorithm that properly de-duplicates through compositions remains to be developed. But after all the filtering listed here, there is usually a quite compact initial set of pooling operations for further optimizations.

The following is a quick lookup table of pooling operations for a few common equivariance types.

**Putting everything together: The equivariant Einsum network.** With an equivariant layer, we can stack them to create a practical high capacity neural net that learns well. Let's apply the following recipe

1) Stacking multiple equivariant layer to create a deep network.

2) GELU nonlinearity between equivariant layers to add to the depth and create bottlenecks.

3) Residual connections for better optimization dynamics.

4) Average pooling to create invariant dimensions if the symmetry involves invariance.

The result is an equivariant backbone as Figure 2.

High order einsum operations can result in large values. One consideration in practice is how to normalize them to be in line with lower order terms. Our observation is that mean subtraction and standard deviation normalization seems to help learning, but further research is still needed to improve learning stability.

Another consideration in practice is einsum path optimization[2]. For example, the einsum string ab,dc,ae,ac,db−>de by default is programmed to be computed pairwise from left to right. By the third term, a large factor abcde would be created and stress the memory. Instead, if we compute pairwise via path ab,db−>ad, ac,dc−>ad, ad,ad−>ad and ad,ae−>de, the largest intermediate factor would only be 2-dimensional and the computation can also be done much faster. For modeling complex higher-order interations under certain types of symmetries, large einsums may be unavoidable, and computing them might be an interesting compute challenge.

Putting it all together, here is a reference Pytorch implementation of an Equivariant Einsum Network that would served as the backbone, to be followed by averaging for dimensions that need invariance.

```python
import torch
import torch.nn as nn
import torch.nn.functional as F

#Implements minimal aH-type pooling
class einpool_a(nn.Module):
    fan_in=4
    fan_out=3
    ndims=1
    def forward(self,x_):
        x=x_.view(-1,*x_.shape[-2:]) # Apply pooling only to the
            last 2 dims, supposedly `aH`
        N,KH=x.shape[-3:]
        H=KH//self.fan_in
        x=x.split(H,dim=-1)
        y0=x[0]
        y1=x[1].mean(-2,keepdim=True).repeat(1,N,1)
        y2=x[2]*x[3]
        y=torch.cat((y0,y1,y2),dim=-1)
        y=y.view(*x_.shape[:-1],-1) #Recover original tensor
            shape
        return y

#Implements minimal aaH-type pooling
class einpool_aa(nn.Module):
    fan_in=8
    fan_out=6
    ndims=2
    def forward(self,x_):
        x=x_.view(-1,*x_.shape[-3:]) # Apply pooling only to the
            last 3 dims, supposedly `aaH`
        N,M,KH=x.shape[-3:]
        H=KH//self.fan_in
        xn=x.view(-1,N*M,KH)
        xn=F.normalize(xn-xn.mean(-2,keepdim=True),
            dim=-2,p=2,eps=1e-1).view(*x.shape)
        x=x.split(H,dim=-1)
        xn=xn.split(H,dim=-1)
        y0=x[0]

            y1=x[1].diagonal(dim1=-2,dim2=-3).diag_embed(dim1=-2,dim2=-3)
```

---

[2]https://numpy.org/doc/stable/reference/generated/numpy.einsum_path.html

```python
            y2=x[2].transpose(-2,-3)
            y3=x[3].mean(-2,keepdim=True).repeat(1,1,M,1)
            y4=x[4]*x[5]
            y5=torch.einsum('ZabH,ZbcH->ZacH',xn[6],x[7])
            y=torch.cat((y0,y1,y2,y3,y4,y5),dim=-1)
            y=y.view(*x_.shape[:-1],-1) #Recover original tensor
             ↪  shape
            return y


#Implements order-3 abH-type pooling
class einpool_ab(nn.Module):
    fan_in=8
    fan_out=5
    ndims=2
    def forward(self,x_):
        x=x_.view(-1,*x_.shape[-3:]) # Apply pooling only to the
         ↪  last 3 dims, supposedly `abH`
        N,M,KH=x.shape[-3:]
        H=KH//self.fan_in
        xn=x.view(-1,N*M,KH)
        xn=F.normalize(xn-xn.mean(-2,keepdim=True),
         ↪   dim=-2,p=2,eps=1e-12).view(*x.shape)
        #xn=F.softmax(x.view(-1,N*M,KH),dim=-2).view(*x.shape)
        x=x.split(H,dim=-1)
        xn=xn.split(H,dim=-1)
        y0=x[0]
        y1=x[1].mean(-2,keepdim=True).repeat(1,1,M,1)
        y2=x[2].mean(-3,keepdim=True).repeat(1,N,1,1)
        y3=xn[3]*x[4]
        y4=torch.einsum('ZacH,ZbcH,ZadH->ZbdH',xn[5],xn[6],x[7])
        y=torch.cat((y0,y1,y2,y3,y4),dim=-1)  #
        y=y.view(*x_.shape[:-1],-1) #Recover original tensor
         ↪  shape
        return y


#Equivariant EinNet layer
class einnet_layer(nn.Module):
    def __init__(self,ninput,nh0,noutput,pool):
        super().__init__()
        self.fan_in=nn.Linear(ninput,nh0*pool.fan_in)
        self.fan_out=nn.Linear(nh0*pool.fan_out,noutput)
        self.pool=pool

    def forward(self,x):
        h=self.fan_in(x)
        h=self.pool(h)
        return self.fan_out(h)

#Equivariant EinNet backbone
class einnet(nn.Module):
    def __init__(self,ninput,nh0,nh,noutput,nstacks,pool):
        super().__init__()
        self.t=nn.ModuleList()
        self.t.append(einnet_layer(ninput,nh0,nh,pool))
        for i in range(nstacks-1):
            self.t.append(einnet_layer(nh,nh0,nh,pool))
```

```python
        self.t.append(einnet_layer(nh,nh0,noutput,pool))
        self.pool=pool

    # Forward call
    #   x: tensor shape matches equivariance type, e.g. *abH
    def forward(self,x):
        h=self.t[0](x)
        for i in range(1,len(self.t)-1):
            h=h+self.t[i](F.gelu(h))

        return self.t[-1](F.gelu(h))

#Example usage
#   net=einnet(1,16,64,1,2,einpool_ab())
```



\end{document}